\documentclass{article}

\usepackage{arxiv}

\usepackage[utf8]{inputenc} 
\usepackage[T1]{fontenc}    
\usepackage{hyperref}       
\usepackage{url}            
\usepackage{booktabs}       
\usepackage{amsfonts}       
\usepackage{nicefrac}       
\usepackage{microtype}      
\usepackage{lipsum}		
\usepackage{graphicx}
\usepackage{doi}
\usepackage{threeparttablex}
\usepackage{tcolorbox}
\usepackage{url}
\usepackage{amsmath}
\usepackage{array}
\newcolumntype{L}[1]{>{\raggedright\let\newline\\\arraybackslash\hspace{0pt}}m{#1}}
\newcolumntype{C}[1]{>{\centering\let\newline\\\arraybackslash\hspace{0pt}}m{#1}}
\newcolumntype{R}[1]{>{\raggedleft\let\newline\\\arraybackslash\hspace{0pt}}m{#1}}
\usepackage{siunitx}
\newcounter{promptcounter}
\newenvironment{promptbox}[1]{%
    \refstepcounter{promptcounter}%
    \begin{tcolorbox}[
        colback=gray!10,
        colframe=gray!90,
        boxrule=1pt,
        title={Prompt \thepromptcounter: #1},
        fonttitle=\bfseries
    ]
}{%
    \end{tcolorbox}
}
\usepackage{import}

\usepackage{amsthm}

\usepackage{booktabs}
\usepackage{amsmath}

\title{Overcoming data challenges through enriched validation and targeted sampling to measure whole-person health in electronic health records}

\title{On Using Large Language Models to Enhance Clinically-Driven Missing Data Recovery Algorithms in Electronic Health Records}

\date{\today} 					

\author{ \href{https://orcid.org/0000-0001-5380-2427}{\includegraphics[scale=0.06]{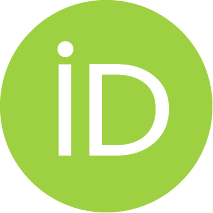}\hspace{1mm}Sarah C.~Lotspeich} \\
	Department of Statistical Sciences\\
	Wake Forest University\\
	Winston-Salem, NC 27109 \\
	\texttt{lotspes@wfu.edu} \\
	\And
    \href{https://orcid.org/0000-0002-7589-7698}{\includegraphics[scale=0.06]{orcid.pdf}\hspace{1mm}Abbey ~Collins} \\  
    Department of Psychology \\
    North Carolina State University \\
    Raleigh, NC 27607 \\
    \And
    \href{https://orcid.org/0000-0001-7310-6525}{\includegraphics[scale=0.06]{orcid.pdf}\hspace{1mm}Brian J. ~Wells} \\  
    Department of Biostatistics and Data Science\\
	Wake Forest University School of Medicine \\
	Winston-Salem, NC 27157 \\
    \And 
    \href{https://orcid.org/0000-0002-9083-891X}{\includegraphics[scale=0.06]{orcid.pdf}\hspace{1mm}Ashish K. ~Khanna} \\ 
    Department of Anesthesiology,\\ Division of Critical Care Medicine, \\
    Wake Forest University School of Medicine \\
	Winston-Salem, NC 27157 \\
    Outcomes Research Consortium\\
	Houston, TX 77030 \\
    \And 
    \href{https://orcid.org/0000-0001-6265-0752}{\includegraphics[scale=0.06]{orcid.pdf}\hspace{1mm}Joseph ~Rigdon} \\ 
    Department of Biostatistics and Data Science\\
	Wake Forest University School of Medicine \\
	Winston-Salem, NC 27157 \\
    \And
    \href{https://orcid.org/0000-0001-5380-2427}{\includegraphics[scale=0.06]{orcid.pdf}\hspace{1mm}Lucy ~D'Agostino McGowan} \\
	Department of Statistical Sciences\\
	Wake Forest University\\
	Winston-Salem, NC 27109 \\
}

\renewcommand{\shorttitle}{On Using LLM to Enhance Clinically-Driven Missing Data Recovery Algorithms in EHR}

\hypersetup{
pdftitle={ALI in EHR with LLM},
pdfsubject={q-bio.NC, q-bio.QM},
pdfauthor={Sarah C.~Lotspeich, Abbey ~Collins, Brian J. ~Wells, Ashish K. ~Khanna, Joseph ~Rigdon, Lucy ~D'Agostino McGowan},
pdfkeywords={chart reviews, computable phenotype, learning health system, missing data, whole-person health},
}

\begin{document}
\maketitle

\begin{abstract}
\textbf{Objective:} Electronic health records (EHR) data are prone to missingness and errors. Previously, we devised an "enriched" chart review protocol where a "roadmap" of auxiliary diagnoses (anchors) was used to recover missing values in EHR data (e.g., a diagnosis of impaired glycemic control might imply that a missing hemoglobin A1c value would be considered unhealthy). Still, chart reviews are expensive and time-intensive, which limits the number of patients whose data can be reviewed. Now, we investigate the accuracy and scalability of a roadmap-driven algorithm, based on ICD-10 codes (International Classification of Diseases, 10th revision), to mimic expert chart reviews and recover missing values.\\ 
\textbf{Materials and Methods:} In addition to the clinicians' original roadmap from our previous work, we consider new versions that were iteratively refined using large language models (LLM) in conjunction with clinical expertise to expand the list of auxiliary diagnoses. Using chart reviews for $100$ patients from the EHR at an extensive learning health system, we examine algorithm performance with different roadmaps. Using the larger study of $1000$ patients, we applied the final algorithm, which used a roadmap with clinician-approved additions from the LLM. \\
\textbf{Results:} The algorithm recovered as much, if not more, missing data as the expert chart reviewers, depending on the roadmap. \\
\textbf{Discussion:} Clinically-driven algorithms (enhanced by LLM) can recover missing EHR data with similar accuracy to chart reviews and can feasibly be applied to large samples. Extending them to monitor other dimensions of data quality (e.g., plausability) is a promising future direction. 
\end{abstract}


\section*{Background and Significance}

\subsection*{Learning health systems and electronic health records}

Refining healthcare systems is vital to enhancing care delivery and improving patient outcomes. Learning health systems (LHS) use data and analytics to learn from patients' data and relay that information back to clinicians, fostering cycles of continuous improvement.\cite{greene_implementing_2012,easterling_clarifying_2022,friedman_toward_2015} Enhancing connections between research and practice is critical to LHS,\cite{easterling_clarifying_2022} and this connection depends heavily on electronic health records (EHR) data.\cite{mclachlan_framework_2019, trinkley_applying_2024}

EHR data present a huge opportunity to operationalize computable phenotypes and advance LHS,\cite{Mo2015} but there are challenges. Because they are routinely collected in a fast-paced clinical environment, EHR data are predisposed to missingness and errors.\cite{mclachlan_framework_2019, hersh_caveats_2013, wells_strategies_2013, kim_evolving_2019, nordo_use_2019, Kim2024} Key predictors for the phenotype may be measured for some, but not all. Because their measurement depends on clinical decision-making (e.g., physician ordering a test), these predictors are likely missing not at random.\cite{Haneuse2021,GETZEN2023104269}

\subsection*{Whole-person health scores}

Our LHS wants to operationalize the allostatic load index (ALI), which is a computable phenotype for allostasis (the body’s ability to adjust to physiological changes and maintain stability).\cite{mcewen_concept_2003, Sterling} The ALI is a modifiable index designed to capture physiological ``wear and tear'' across multiple body systems due to stress.\cite{mcewen_stress_1993, mcewen_concept_2003, mcewen_what_2010} It can provide clinicians with an early indication of short and long-term health risk, as the ALI is associated with outcomes including  cardiovascular disease, hypertension, diabetes, obesity, and overall risk of morbidity and mortality. \cite{beckie_systematic_2012, duru_allostatic_2012, gruenewald_combinations_2006, guidi_allostatic_2020, guidi_allostatic_2020-1, mcewen_stress_1998, parker_allostatic_2022, Seeman2001} 

We want to apply the calculation from Seeman et al.,\cite{Seeman2001} which computes ALI from ten biomarkers across three body systems (Table~\ref{tab:ali_components_roadmap}). However, in EHR data, many of these biomarkers, especially the rare inflammatory labs, are measured infrequently. While errors in non-missing biomarkers are possible, based on our previous studies in this LHS,\cite{lotspeich2025overcomingdatachallengesenriched} we believe missingness to be a bigger issue. We recently assessed how to navigate data challenges and more accurately measure ALI from EHR data by incorporating expert chart reviews.\cite{lotspeich2025overcomingdatachallengesenriched}  

\begin{table}[ht!]
\centering
\resizebox{\columnwidth}{!}{
\begin{tabular}{rcc}
\hline 
        \textbf{ALI Component}& \textbf{Threshold} & \textbf{Search Term(s)} \\
        \hline 
        \addlinespace
        \multicolumn{3}{l}{\textit{Cardiovascular System}} \\
        \addlinespace
        Systolic Blood Pressure & $>140$ & Hypertension  \\
        Diastolic Blood Pressure & $>90$ & Hypertension  \\
        \addlinespace
        \multicolumn{3}{l}{\textit{Metabolic System}} \\
        \addlinespace
        Body Mass Index & $>30$ & Obesity; Morbid Obesity; \\
        & & Grade I, II, or III Obesity\\
        Triglycerides & $\geq 150$ &	Hypertriglyceridemia \\
        Total Cholesterol & $\geq 200$ & Hypercholesterolemia \\
        \addlinespace
        \multicolumn{3}{l}{\textit{Inflammation System}} \\
        \addlinespace
        C-Reactive Protein & $\geq 10$	& Sepsis; Infection; Auto-Immune; Inflammatory Syndrome\\
        Hemoglobin A1C & $\geq 6.5$	& Diabetes; Impaired \\
        & & Glycemic Control \\
        Serum Albumin & $\geq 3.5$ & \textit{(None Given)} \\
        Creatinine Clearance & $<110$ (Males)& Renal Failure; Insufficiency; Acute Kidney Injury; Chronic Renal Failure\\
        & $<100$ (Females)  & \\
        Homocysteine & $>50$ &	Hyperhomocysteinemia; Vitamin deficiency \\
        \hline 
    \end{tabular}
}
\caption{Ten components of the allostatic load index (ALI) were defined by discretizing measurements across three body systems at clinically-driven thresholds. If the component was missing from the extracted EHR data, chart reviewers searched for auxiliary information in the patient's medical chart in Epic (the institution's electronic charting software). If the search term was present, the patient was treated as having an ``unhealthy'' measurement. This table was adapted from Lotspeich et al. (2025). \cite{lotspeich2025overcomingdatachallengesenriched}}
\label{tab:ali_components_roadmap}
\end{table}

\subsection*{Expert chart reviews}

Expert chart reviews (also called source document verification) are a common approach to gauging the quality of routinely collected data.\cite{kiragga2011, mphatswe2012, chaulagai2005, kimaro2005, shepherd2022, lotspeich2025overcomingdatachallengesenriched, duda2012} Typically, teams of skilled reviewers manually compare the analytical dataset (e.g., extracted from the EHR) to clinical source documents (e.g., electronic charts) to identify differences. Since the source documents are closer to the point of care, they are generally considered more accurate than the analytical dataset. With a prespecified protocol in place,\cite{duda2012,Lotspeich2020,Lotspeich2023} chart reviews can take weeks or months to complete (depending on the number of patients and variables), and data are often collected in a standardized electronic format, like Microsoft Excel or REDCap.\cite{Harris2009}

Chart reviews uncover valuable information about data quality and opportunities for process improvement. Still, they are resource- and time-intensive, which greatly limits their scalability. Our previous chart reviews required the effort of four skilled chart reviewers (clinical research trainees) over $6$ months to evaluate a focused list of variables (only ALI components) for a small subsample of $100$ patients from a larger study of $1000$ routine healthcare users from our EHR. Adding a novel ``roadmap'' for missing data recovery, created by clinical experts, we filled in some gaps using auxiliary health information (e.g., related diagnoses) or anchors,\cite{Halpern2014} but $900$ patients' data were left totally untouched, and even those who underwent chart review could still have missing data. Herein, we explore how (i) expanding the roadmap, potentially with large language models (LLM), and (ii) applying it algorithmically, instead of requiring humans for implementation, can promote this protocol's efficacy and efficiency. 

\subsection*{Large-scale algorithms in electronic health records}

Structured EHR data, like ICD-10 codes (International Classification of Diseases, 10th revision),\cite{WHO_ICD10} are especially valuable for patient phenotyping.\cite{Pendergrass2019,Zhang2019,Yang2023,Petrik2016,Yang2025, LEDERMACEK2021101599,LIU2023105136,He2024,Barnes2020,Stein2019,JORGE201984,Yuan2021} Often, phenotype development involves model training and deployment using EHR data and a subset of ``gold standard" outcomes (e.g., validated diagnoses) obtained through chart reviews.\cite{JORGE201984} The final algorithm might involve statistical modeling, machine learning, or artificial intelligence. 

Data cleaning algorithms differentiate between plausible and implausible values in EHR data,\cite{Dziadkowiec2016,Estiri2019,Shi2021} and they can also detect misspellings in text fields.\cite{Filik2006,Shi2021,LAI2015188, Lee2022} Some algorithms are variable-specific, since they rely on the clinical context to define plausibility (e.g., when identifying erroneous heights and weights).\cite{Daymont2017, Lin2022, GUIDE2024104660} 

To handle missing data in EHR, imputation (i.e., replacing missing values with informative placeholders), which seeks to preserve the size and statistical power of the full sample, is incredibly common.\cite{Wells2013,Wang2014,Goldstein2016,Beaulieu-Jones2018,Xu2021,BERNARDINI2023107188,Jazayeri2020} The imputed values (placeholders) range in complexity from summary statistics to model predictions. However, replacing missing values with poor placeholders can actually hurt rather than help analyses.\cite{Li2018} To recover missing ALI components, we did not want to rely on relationships between variables in our somewhat sparse EHR dataset. Instead, we leverage clinical insights to impute variables based on a prespecified roadmap. 

\section*{Objective}

We compare the accuracy and scalability of a missing data recovery algorithm based on a clinically-driven, LLM-enhanced roadmap of auxiliary ICD-10 codes against expert chart reviews. 

\section*{Materials and Methods}
\subsection*{Cohort description}

Our study builds on a sample of $N = 1000$ adults $18$--$65$ years old who newly engaged in outpatient primary care at Atrium Health Wake Forest Baptist Hospital between March $2018$--$2020$. Located in Winston-Salem, North Carolina, this hospital belongs to a statewide LHS. This study was approved by the Institutional Review Board at Wake Forest University School of Medicine.

EHR data were extracted for all $1000$ patients, including demographics and  measurements for the ten ALI components (Table~\ref{tab:ali_components_roadmap}). 
Some components were missing for almost everyone (Figure~S1A). Errors in non-missing values were possible (e.g., due to the data extraction algorithm) but very uncommon. On average, patients had $6$ of $10$ non-missing ALI components in the EHR data (interquartile range [IQR] $= [5, 7]$, Figure~S1B). A subset of $n = 100$ patients underwent chart reviews.\cite{lotspeich2025overcomingdatachallengesenriched} 

\subsection*{Expert chart review protocol and findings}

We previously developed a novel ``enriched'' chart review protocol incorporating a clinical expert-derived ``roadmap" (Table~\ref{tab:ali_components_roadmap}) to ensure the quality and completeness of ALI components in EHR data. \cite{lotspeich2025overcomingdatachallengesenriched} It was designed to address two key goals. 
\begin{enumerate}
\item[(i)] \textit{Validation:} For non-missing components, confirm that the extracted value (in extracted EHR data) matches the patient's chart.
\item[(ii)] \textit{Recovery:} For missing components, locate supplemental diagnoses (in electronic charts) that provide information about the unmeasured values.   
\end{enumerate}
These goals aligned with two data quality dimensions (concordance and completeness) outlined by Weiskopf et al. (2013).\cite{Weiskopf2013} 

Expert chart reviewers (clinical research technicians [CRTs]) followed this protocol as they manually searched patients' electronic charts. In total, four CRTs reviewed \num{11472} data points ($1767$ labs and $9705$ vitals). These data are ``alloyed'' gold standards (i.e., presumed to be more accurate than extracted EHR data but still imperfect). Non-missing ALI components almost always matched the EHR, and some missing ones were recovered using the roadmap. Still, chart reviews required $100$ hours of human effort. Also, while we now feel confident in the non-missing EHR data, missingness remains a key challenge (especially for the $900$ patients without chart reviews). 

\subsection*{Missing data recovery algorithms}

Since completing the chart reviews, we have been pursuing scalable ways to recover missing EHR data across our healthcare system. The clinicians' search terms from the roadmap in our previous work (Table~\ref{tab:ali_components_roadmap}) could be captured by International Classification of Diseases (ICD-10) codes (10th revision).\cite{WHO_ICD10} Therefore, we were interested in the possibility of developing an algorithm  to recover missing ALI components from all $1000$ patients on the basis of their ICD-10 codes. Computationally, this task is fairly simple: Merge the roadmap into patient diagnoses and evaluate potential matches. 

Our merging algorithm required all terms to appear somewhere in the ICD-10 code's description to be considered a match. For example, ``vitamin deficiency'' required both ``vitamin'' and ``deficiency" to be present. The $20$ search terms in the \textit{clinicians' original roadmap} matched $1234$ ICD-10 codes, of which $211$ were present in our sample (Table~\ref{tab:terms_codes}). 

\begin{table}[ht!]
\centering
\resizebox{\columnwidth}{!}{
\begin{tabular}{p{1.5cm}p{1.5cm}p{1.5cm}p{1cm}p{1.25cm}p{1.5cm}p{1.5cm}p{1.5cm}p{2cm}} 
  \hline
  \textbf{Data Source} & \textbf{Addresses Errors} & \textbf{Addresses Missingness} & \textbf{Sample Size} & \textbf{List of Diagnoses} & \textbf{Clinical Relevance} & \textbf{Statisti- cally Efficient} & \textbf{Search Terms} & \textbf{Matched in Sample (Overall)}\\
  \hline \\
  Extracted EHR Data & No & No & Large & Short & High & Maybe & $-$ & $-$   \\ \\
  Expert Review Reviews & Yes & Yes & Small & Short & High & No & $-$ & $-$ \\ \\
  Algorithm w/ Clinicians' Original & No & Yes & Large & Short & High & Yes & $20$ & $211$ ($1234$) \\ \\
  Algorithm w/ LLM (Baseline) & No & Yes & Large & Long & Low & Yes & $656$ & $118$ ($539$) \\ \\
  Algorithm w/ LLM (Context) & No & Yes & Large & Long & Low & Yes & $950$ & $275$ ($1853$) \\ \\
  Algorithm w/ LLM (Context + Clinicians) & No & Yes & Large & Long & High & Yes & $79$ & $243$ ($243^*$) \\ \\
  \hline 
\end{tabular}}
\caption{Comparison of the pros and cons for various data sources, including the extracted electronic health records (EHR) data, expert chart reviews, and various missing data recovery algorithms (by roadmap). For each roadmap, counts of proposed search terms, matching International Classification of Disease (ICD-10) codes (10th revision), and matching ICD-10 codes among patients in our sample. $^*$Clinicians were only asked to review ICD-10 codes from LLM (context) that matched patients in our sample; it was not possible for LLM (context + clinicians) to match more than $275$ ICD-10 codes.}
\label{tab:terms_codes}
\end{table}

This algorithm attempts recovery missing data by replicating the process performed by human chart reviewers. Once established, it is faster, more affordable, and efficiently applies across the entire study. Since costs are not tied to sample size, this algorithm makes large-scale EHR data recovery possible in a way that chart reviews cannot. 

However, the roadmap was initially created for use by human chart reviewers with clinical expertise. Therefore, some search terms were expected to be used at the chart reviewers' discretion based on additional context. For example, one of the search terms for missing C-reactive protein (CRP) was ``infection,'' which matched ICD-10 codes ranging in severity from systemic (like sepsis) to acute (like upper respiratory infections).

\subsection*{Enhancements with large language models (LLM)}

Building on this algorithmic foundation, we used LLM to expand the list of search terms. We implemented this \textit{LLM-based roadmap enhancement} using the \texttt{ellmer} package in \texttt{R}, which enables programmatic interaction with LLM through ``tool-calling.''\cite{ellmer} That is, the \texttt{ellmer} package allows LLM to request execution of \texttt{R} functions as external tools.  Using Gemini-2.5-Flash,\cite{geminiteam2023gemini} we tested two LLM enhancements. See the Supplemental Materials for all \texttt{R} code.

\subsubsection*{LLM (baseline) roadmap without context} 

First, we prompted the LLM to generate relevant terms for each ALI component based only on the name of each biomarker and general guidance on whether high or lower values were considered unhealthy (Prompt~\ref{prompt:generation}). 
\begin{promptbox}{Generate LLM (baseline) roadmap}
\label{prompt:generation}
Please propose an exhaustive list of terms (avoiding acronyms) that will be used to search ICD descriptions to identify each of the missing biomarkers and create a dataframe with these codes. I want you to repeat this process 20 times, creating a new dataframe each time with each having a unique name starting with \texttt{df\_nocontext}. Each time you repeat this, be sure to make as exhaustive a list as possible. These lists can vary.
\end{promptbox}
\vspace{-5pt}
\begin{small}
\end{small}
\vspace{5pt}

Ultimately, the \textit{LLM (baseline) roadmap} took the superset of these $20$ dataframes (i.e., the collection of terms that appeared at least once across the $20$ times that the LLM completed the task). The LLM (baseline) proposed $656$ search terms, not necessarily including the original $20$ from the clinicians' original roadmap. While $539$ of them matched potential ICD-10 codes, only $118$ actually matched diagnoses for patients in our sample. 
Notably, despite starting with $33$ times as many terms, the LLM (baseline) roadmap matched essentially half as many patient diagnoses as the clinicians' original (Table~\ref{tab:terms_codes}). 

\subsubsection*{LLM (context) roadmap} 

Second, building on the same function for the roadmap's structure, the prompt was modified slightly to instruct the LLM to include the examples from the clinicians' original roadmap (Table~\ref{tab:ali_components_roadmap}) in each iteration (Prompt~\ref{prompt:generation-context}). Again, the \textit{LLM (context) roadmap} comprised all unique search terms across the $20$ resulting dataframes. 
\begin{promptbox}{Generate LLM (context) roadmap}
\label{prompt:generation-context}
Please propose an exhaustive list of terms (avoiding acronyms) that will be used to
search ICD descriptions to identify each of the missing biomarkers and create a dataframe with these codes. I want you to repeat this process 20 times, creating a new dataframe each time with each having a unique name starting with \texttt{df\_context}. Each time you
repeat this, be sure to include the examples given in (e.g.,) and make as exhaustive a
list as possible. These lists can vary.
\end{promptbox}
\vspace{-5pt}
\begin{small}
\end{small}
\vspace{5pt}
LLM (context) proposed $950$ search terms, which matched $1853$ distinct ICD-10 codes ($275$ in our sample). Not only did LLM (context) identify more initial terms than LLM (baseline), but the suggested terms matched $2.3$--$3.4$ times as many valid ICD-10 codes (overall and in our sample). They also matched $1.2$-$1.5$ times as many as the clinicians' original roadmap (Table~\ref{tab:terms_codes}). 

\subsection*{Clinicians' adjudication on the LLM (context) roadmap}

We considered one final roadmap, where two clinicians independently reviewed the proposed LLM (context) additions and adjudicated whether the search terms were valid for the corresponding ALI components. We focused on the LLM (context), which was expected to outperform the LLM (baseline).\cite{Groza2024} Of the $275$ ICD-10 codes matching patients in our sample based on LLM (context), at least one clinician endorsed $243$ of them ($88\%$). 

Some disqualified terms were actually matches from the original clinicians' roadmap, like ``diabetes inspidus'' or family histories of matching diseases. However, some exclusions, like ``pulmonary hypertension'' were new additions by the LLM (context) that clinicians did not find clinically relevant for the corresponding ALI component. Still, the \textit{LLM (context + clinicians) roadmap} matched slightly more ($1.15$ times as many) patient diagnoses in our sample than the original. 


\section*{Results}

\subsection*{Descriptive statistics}

The sample of $N = 1000$ patients was $48$ years old, on average, $40\%$ male, $72\%$ White, $18\%$ Black, and $93\%$ not Hispanic or Latino (Table~\ref{tab:cohort_desc}). The subset of $n = 100$ patients with chart reviews was slightly older (median age of $50$ years) with higher percentages male ($48\%$), White ($74\%$), Black ($20\%$), and not Hispanic or Latino ($95\%$). 

\begin{table}[ht!]
\centering
\begin{tabular}{R{2.5cm}ccc}
\hline
& & \multicolumn{2}{c}{\textbf{Expert Chart Review?}} \\
\cmidrule(l{3pt}r{3pt}){3-4} 
& \textbf{Overall} & \textbf{No} & \textbf{Yes} \\
& ($\boldsymbol{N = 1000}$) & ($\boldsymbol{N -n = 900}$) & ($\boldsymbol{n = 100}$)\\
\hline
\addlinespace
\multicolumn{4}{l}{\textit{Patient Demographics}} \\
\addlinespace
\multicolumn{1}{l}{Age} & $48$ ($35$, $57$) & $48$ ($35$, $57$) & $50$ ($31$, $59$)\\
\multicolumn{1}{l}{Sex} \\
Male (\%) & $395$ ($39.5$) & $347$ ($38.6$) & $48$ ($48.0$)\\
\multicolumn{1}{l}{Race} \\
\hspace{2mm}American Indian or Alaska Native
& $6$ ($0.6$) & $5$ ($0.6$) & $1$ ($1.0$)\\
Asian Indian & $30$ ($3.0$) & $29$ ($3.2$) & $1$ ($1.0$)\\
Black or African American & $181$ ($18.1$) & $161$ ($17.9$) & $20$ ($20.0$)\\
Other & $68$ ($6.8$) & $64$ ($7.1$) & $4$ ($4.0$)\\
White or Caucasian & $715$ ($71.5$) & $641$ ($71.2$) & $74$ ($74.0$)\\
\multicolumn{1}{l}{Ethnicity} \\
Hispanic, Latino or Spanish & $61$ ($6.1$) & $56$ ($6.2$) & $5$ ($5.0$)\\
Not Hispanic, Latino or Spanish & $934$ ($93.4$) & $839$ ($93.2$) & $95$ ($95.0$)\\
Patient Refused & $5$ ($0.5$) & $5$ ($0.6$) & $0$ ($0.0$)\\
\addlinespace
\multicolumn{4}{l}{\textit{Allostatic Load Index (ALI) and Healthcare Utilization}} \\
\addlinespace
\multicolumn{1}{L{3cm}}{Unvalidated ALI}
& $0.33$ ($0.17$, $0.50$) & $0.33$ ($0.17$, $0.50$) & $0.38$ ($0.17$, $0.57$)\\
\multicolumn{1}{L{3cm}}{Missing ALI Components}
& $4$ ($3$, $4$) & $4$ ($3$, $4$) & $4$ ($3$, $5$)\\
\multicolumn{1}{L{3cm}}{Unique Diagnosis Codes} & $30$ ($16$, $48$) & $30$ ($16$, $48$) & $32$ ($17$, $47$)\\
\multicolumn{1}{L{3cm}}{Engaged in Care} & $318$ ($31.8$) & $268$ ($29.8$) & $50$ ($50.0$)   \\
\hline
\end{tabular}
\caption{Description of the original sample of $N = 1000$ patients from the electronic health records (EHR) at Atrium Health Wake Forest Baptist Hospital, further broken down by whether or not the patients underwent expert chart review. Categorical variables are summarized by their frequency ($\%$). Numeric variables are summarized by their median (interquartile range). Unvalidated ALI and the number of missing components were taken from the extracted EHR data. Diagnosis codes were based on the ICD-10 (International Classification of Diseases, 10th revision). Patients with $\geq 1$ emergency department visit or hospitalization were considered to have ``engaged in care.''}
\label{tab:cohort_desc}
\end{table}

According to extracted EHR data, the median ALI for patients undergoing chart review was slightly higher than that for the entire sample ($0.38$ versus $0.33$). By design, a much higher proportion of patients undergoing chart review engaged in the healthcare system ($50\%$ versus $32\%$), meaning that they had at least one hospitalization or emergency department (ED) visit during the $2$-year study period. (We intentionally sampled a $50/50$ ratio of patients engaging/not engaging in the healthcare system for chart reviews.\cite{lotspeich2025overcomingdatachallengesenriched}) Patients were missing $4$ ALI components, on average, whether or not they underwent chart review; however, those undergoing chart review had a slightly wider IQR of [$3$, $5$] versus [$3$, $4$]. On average, there were $30$ unique ICD-10 codes per patient in the overall sample (IQR $= [16, 48]$) but $32$ each (IQR $= [17, 47]$) among those with chart reviews. 

We first compared the algorithm's findings, based on various roadmaps, for the $100$ patients who underwent chart reviews. We were interested in (i) how well the ALI components based on the algorithm agreed with those from the chart review, (ii) why the algorithm might recover information that the chart review did not (and vice versa). Then, we scaled the algorithm up for use in the entire sample of $1000$ patients. 

\subsection*{Expert chart reviews versus missing data recovery algorithms}

A total of $1000$ data points were considered in the chart reviews ($10$ ALI components for each of $n = 100$ patients). Each data point was assigned to one of four mutually exclusive categories: unhealthy, healthy, missing, or ``protocol error.'' Data points classified as protocol errors were missing in the extracted EHR data, but, instead of following the roadmap, the chart reviewers found non-missing values that were outside the study period. These data points were returned to missing for our comparisons, while the algorithms might recover them.

\begin{figure*}
    \centering
    \setlength{\abovecaptionskip}{0pt}
    \includegraphics[width=\textwidth]{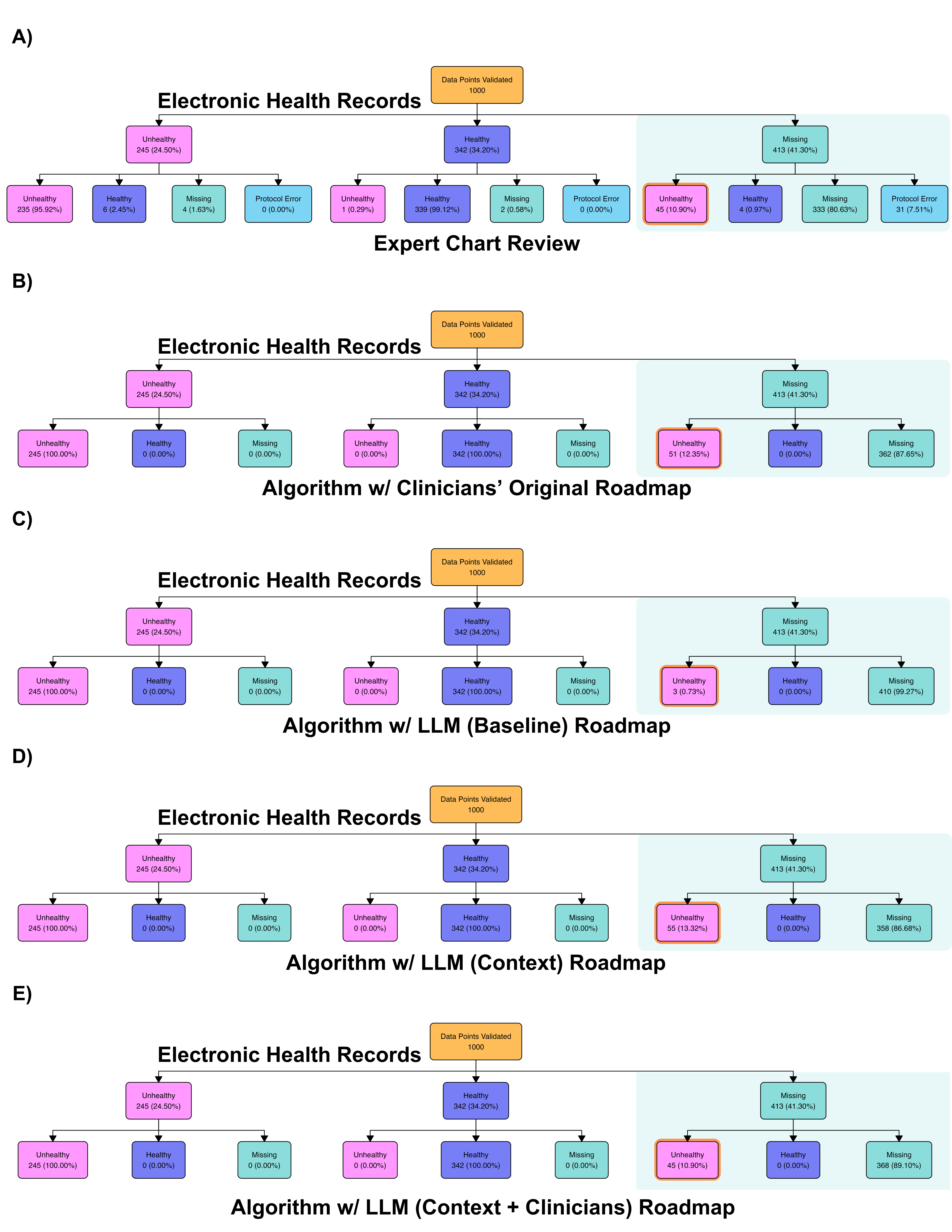}
    \caption{Flow charts of the $1000$ data points included in the expert chart reviews ($10$ components per patient across $n = 100$ patients). We compare the extracted electronic health records (EHR) data to the \textbf{A)} expert chart reviews and the missing data recovery algorithms based on the following roadmaps: \textbf{B)} clinicians' original, \textbf{C)} LLM (baseline), \textbf{D)} LLM (context), and \textbf{E)} LLM (context + clinicians). Shaded boxes denote data points that were missing in the EHR data; these are the only data points that could potentially be changed by the missing data recovery algorithms in \textbf{B}--\textbf{E}, and a missing data point could only be changed to ``unhealthy.''}
    \label{fig:flowcharts}
\end{figure*}

Across all patients and components, $413$ data points were missing in the extracted EHR data. The chart reviews recovered $45$ ($11\%$) that were presumed to be unhealthy (Figure~\ref{fig:flowcharts}A). (After accounting for protocol errors, this value differs slightly from our previous work\cite{lotspeich2025overcomingdatachallengesenriched} but allows for a more fair comparison to the algorithms.) The amount recovered using the algorithm depended heavily on the roadmap (Figure~\ref{fig:flowcharts}B--\ref{fig:flowcharts}E). The clinicians' original recovered slightly more than the chart reviews ($51$, $12\%$). LLM (baseline) only recovered three missing components ($7\%$), while LLM (context) recovered the most ($55$, $13\%$). Finally, LLM (context + clinicians) recovered the same amount as the experts ($45$, $11\%$). 

\begin{figure*}
    \centering
    \setlength{\abovecaptionskip}{0pt}
    \includegraphics[width=\textwidth]{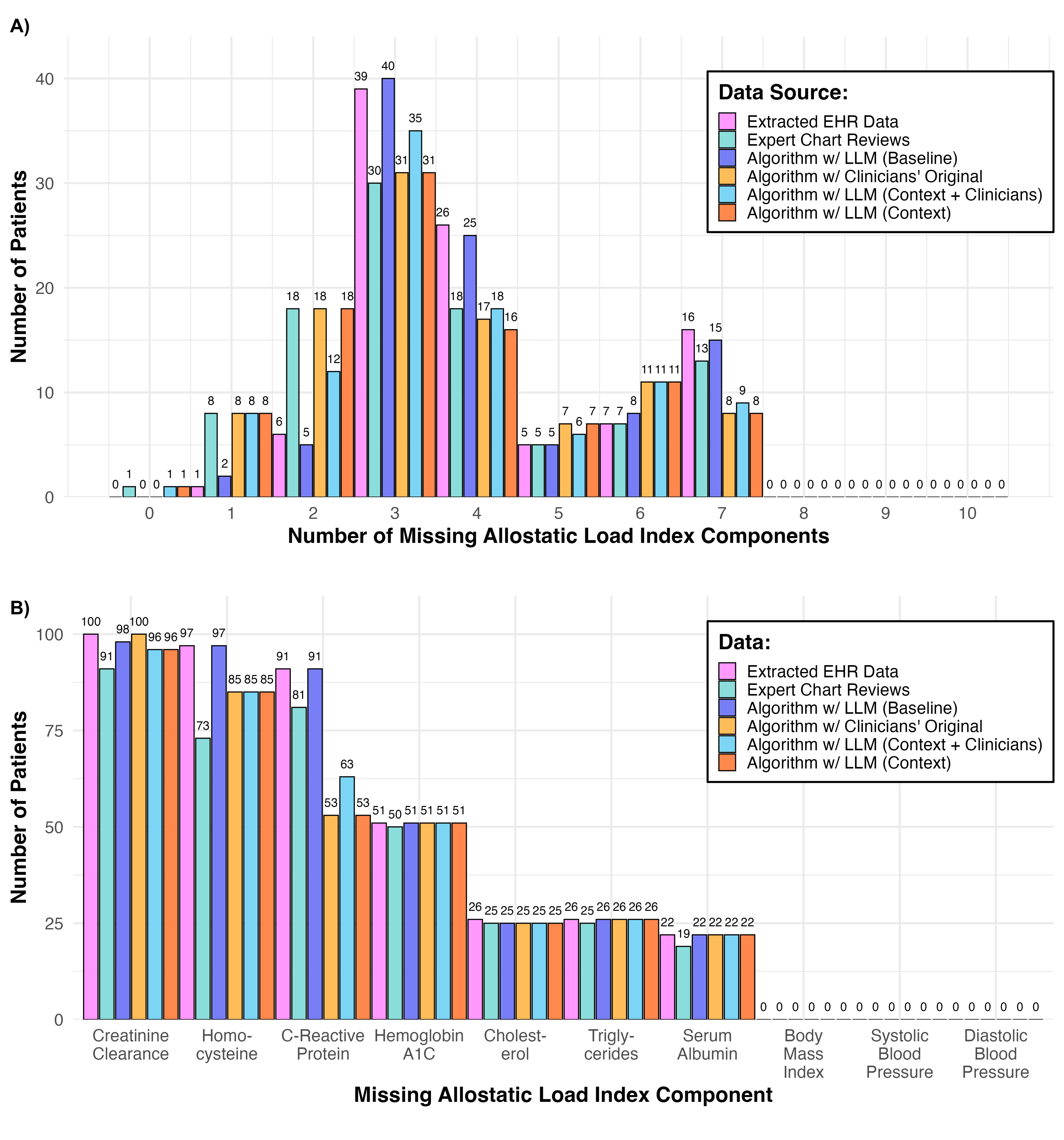}
    \caption{Counts of missing allostatic load index components \textbf{A)} per component and \textbf{B)} per patient across those chosen for expert chart reviews ($n = 100$ patients). Missingness should be highest according to the unvalidated electronic health records (EHR) data, while the chart review might reduce it. The different missing recovery algorithms reduced the amount of missingness by varying amounts, depending on the roadmap: LLM (baseline), clinicians' original, LLM (context + clinicians), or LLM (context).}
    \label{fig:bars_chart_reviews}
\end{figure*}

The counts of missing components per patient can be found in Figure~\ref{fig:bars_chart_reviews}A. In the extracted EHR data, patients had $6$ out of $10$ non-missing ALI components, on average (IQR $= [5, 7]$). Chart reviews improved this average to $7$ out of $10$ (IQR $= [5.75, 8]$). LLM (baseline) roadmap did not improve the per-patient median number of non-missing ALI components (median $= 6$, IQR $= [5, 7]$), but the clinicians' original, LLM (context), or LLM (context + clinicians) all increased it to $7$ out of $10$. The IQR for LLM (context + clinicians) was narrower [$5$, $7$] than that for the  clinicians' original or LLM (context) (both [$5, 8$]).

The counts of missing values per component can be found in Figure~\ref{fig:bars_chart_reviews}B. No body mass index (BMI) or systolic/diastolic blood pressure values were missing. The chart reviews and recovery algorithms all recovered one missing cholesterol component. Across the other six ALI components, two types of disagreement between the chart reviews and recovery algorithms were possible. 

First, the chart reviews could find something that the recovery algorithm did not. Fortunately, there were relatively few of these. For hemoglobin A1C (HbA1c) and triglycerides, the clinical experts recovered one patient's missing component that the algorithms did not. The chart reviewers recovered more patients' homocysteine ($24$) and creatinine clearance ($9$), although these differences depended some on the roadmap. They likely had access to additional information in the electronic chart (e.g., lab values from previous providers). Similarly, no search terms were given for serum albumin, and yet the chart reviewers recovered three missing components.

Second, the algorithm could recover something that the chart reviews did not. We only observed this type of disagreement for CRP, where the clinicians' original and LLM (context) roadmaps seemingly recovered more missing ($38$ versus $10$). These disagreements seemed to stem from clinical interpretation. ``Infection'' was a search term for CRP, and sometimes a matching diagnosis code was present in a patient's chart but deemed not relevant by the chart reviewers (e.g., urinary tract infection). The LLM (context + clinicians) roadmap excluded these diagnoses, and thus the number of CRP values recovered was slightly closer to that from the chart reviews ($28$ versus $10$). 

In general, the LLM (context) roadmap recovered the most missing components ($\leq 38$), followed by the clinicians' original ($\leq 38$), and LLM (context + clinicians) ($\leq 28$). LLM (baseline) recovered very little (just $2$ patients' missing CRP). While the clinicians' original and LLM (context) tended to recover the most, we believe that some of these matches were mistaken and adopted the LLM (context + clinicians) roadmap going forward. 

\subsection*{Scaling up to the entire study} 

We applied the missing data recovery algorithm using the LLM (context + clinicians) roadmap to the entire $1000$-person sample. In total, there were $4009$ missing ALI components, $531$ ($13\%$) of which were recovered by the algorithm. The median ALI for the study was slightly higher after applying it ($0.375$ versus $0.333$ in the extracted EHR data). The distribution of the ALI after recovery was also slightly more symmetric (Supplemental Figure~S2) and less dispersed (IQR $= [0.25, 0.50]$ versus $[0.17, 0.50]$). 

\begin{figure*}
    \centering
    \setlength{\abovecaptionskip}{0pt}
    \includegraphics[width=\textwidth]{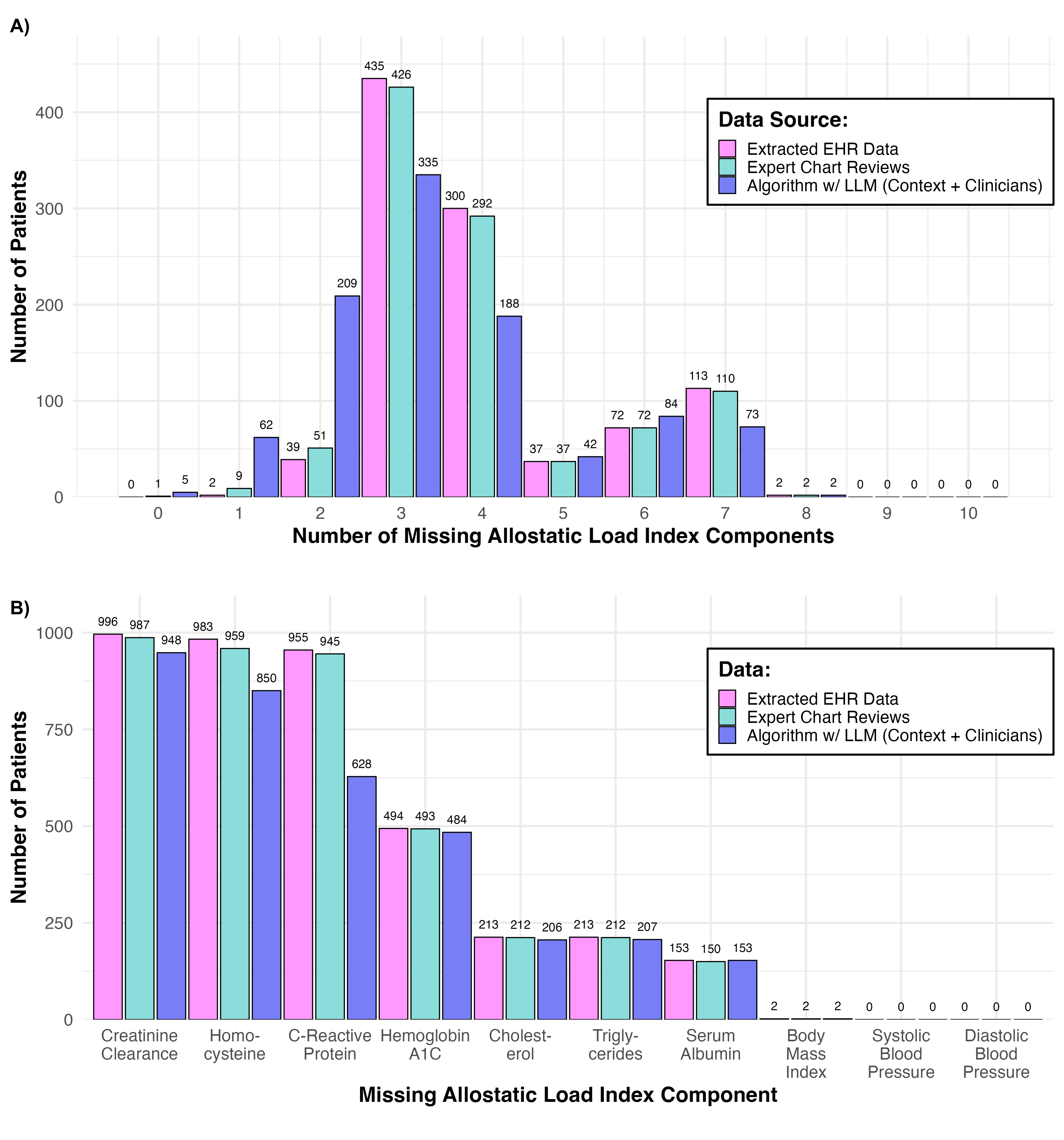}
    \caption{Counts of missing allostatic load index components \textbf{A)} per patient and \textbf{B)} per component across the full sample ($N = 1000$ patients). Missingness should be highest according to the unvalidated electronic health records (EHR) data, while the chart review might reduce it. We only considered algorithmic recovery based on the LLM (context + clinicians) roadmap, which could be applied across the full sample and sometimes recovered more missing data than the expert chart reviews.}
    \label{fig::miss_full_sample}
\end{figure*}

The chart reviews did not move the sample-wide number of non-missing components per patient (median $=6$ with IQR $= [6, 7]$, as in the extracted EHR data), since they only helped $100$ patients. Meanwhile, algorithmic recovery led to $7$ non-missing ALI components per patient, on average (IQR $= [6, 8]$), because the entire $1000$-patient was eligible (Figure~\ref{fig::miss_full_sample}A). 

The algorithm recovered missing values for six of ten ALI components (Figure~\ref{fig::miss_full_sample}B). Noticeably, the most recovered components were for CRP ($327$ of $955$, $34\%$), followed by homocysteine ($133$ of $983$, $14\%$). Small percentages ($\leq 5\%$) of missing creatinine clearance, HbA1c, cholesterol, and triglycerides components were also recovered. Again, no systolic or diastolic blood pressure values were missing. The only components with missing data that the algorithm did not help with were serum albumin (for which there were no search terms) and body mass index (which was only missing for two patients). 

\begin{figure*}
    \centering
    \setlength{\abovecaptionskip}{0pt}
    \includegraphics[width=\textwidth]{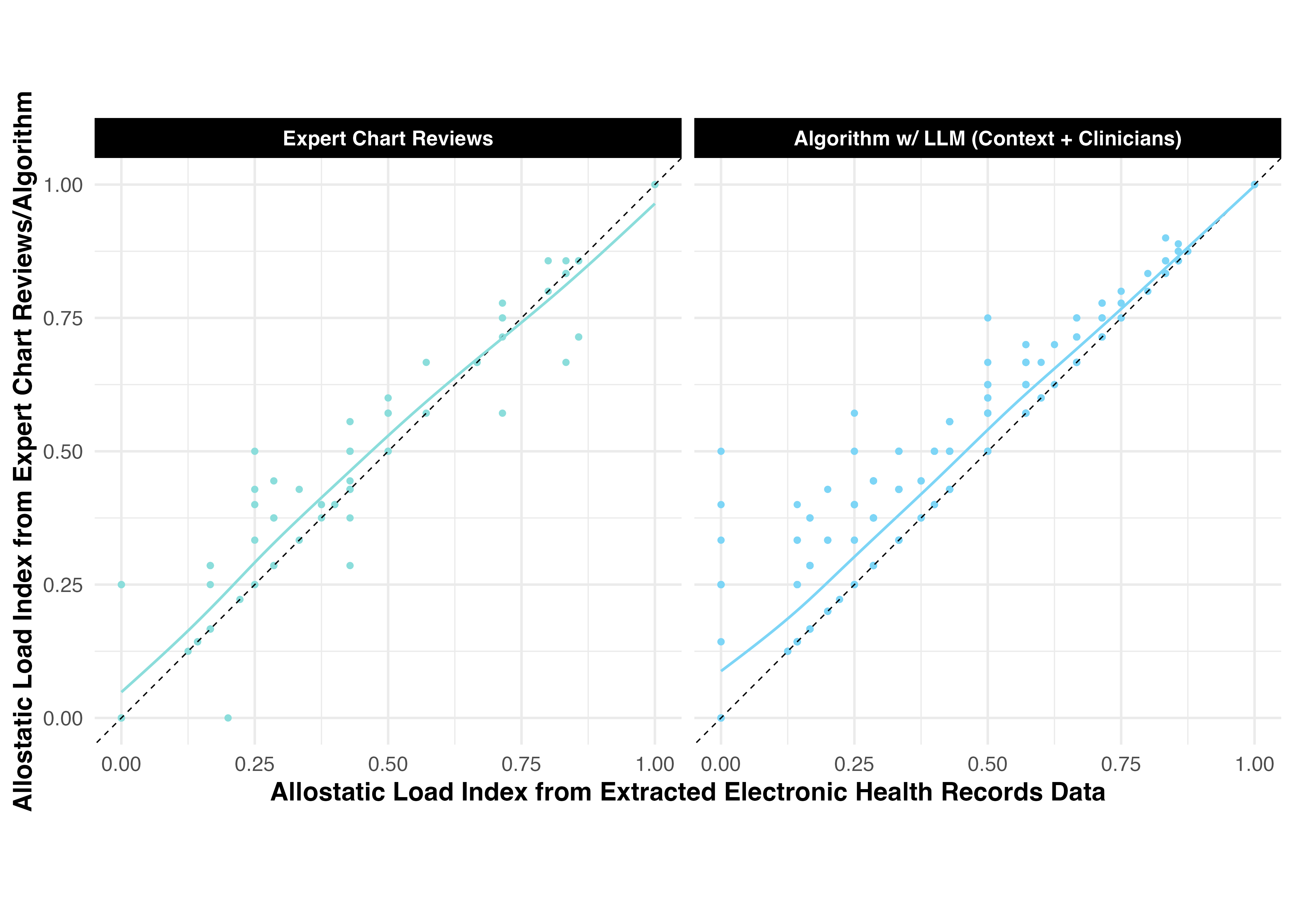}
    \caption{Many patients' expert chart reviewed and algorithmically augmented allostatic load indices (ALIs) differed from the version in their extracted electronic health records (EHR) data. Still, the loess smoothers (solid lines) capturing the relationship between the validated/augmented and unvalidated ALIs were relatively linear and fell close to the line of equality (i.e., where unvalidated and validated/augmented ALI were equal).}
    \label{fig::ali}
\end{figure*}

The association between the ALI from the extracted EHR data versus the recovery algorithm was linear (Figure~\ref{fig::ali}). This relationship, which could be captured with a statistical model, aligned closely with that between ALI from the expert chart reviews and extracted EHR data. If we wanted to attempt to predict a more accurate version of the ALI from EHR data, algorithmic recovery leads to similar values, on average, as having experts manually review a subset of patients. 

\begin{figure*}
    \centering
    \setlength{\abovecaptionskip}{0pt}
    \includegraphics[width=\textwidth]{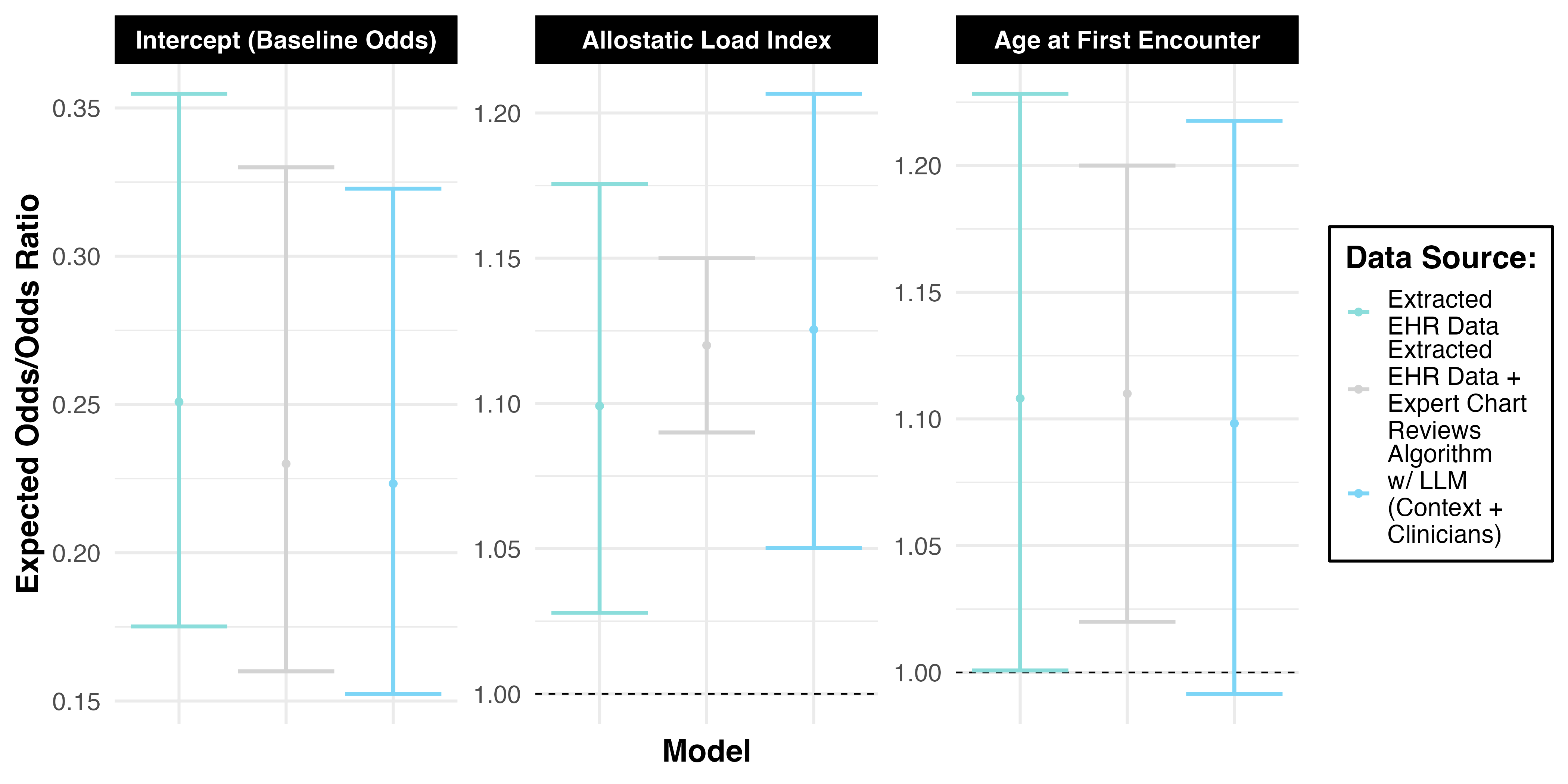}
    \caption{Coefficient estimates (95\% confidence intervals) using the extracted electronic health records (EHR) data (naive analysis), combined EHR data + expert chart reviews (previous study estimates),\cite{lotspeich2025overcomingdatachallengesenriched} and the missing data recovery algorithm using the LLM (context + clinicians) roadmap (augmented analysis).}
    \label{fig::forest}
\end{figure*}

Previously, we found ALI to be strongly associated with odds of engaging in care, after incorporating chart review and EHR data.\cite{lotspeich2025overcomingdatachallengesenriched} (ALI from chart reviews was used in this model when available; otherwise, the reviewed version was essentially predicted from the EHR data.) Here, we wanted to determine whether whole-sample algorithmic recovery could lead us to draw similar conclusions to the subsample of chart reviews. Using standard logistic regression on the $1000$ patients, 
we re-estimated this model using only (i) ALI from the extracted EHR data (``naive analysis'') or (ii) ALI from the algorithm (``augmented analysis''). Overall, the augmented analysis captured many of the same benefits as our incorporation of the chart reviews previously (Figure~\ref{fig::forest}). Namely, the estimated odds and odds ratios (OR) were closer than those from the naive analysis. Our previous estimated had much better precision (i.e., smaller variability); however, it was specifically designed to do so by targeting the most informative patients for chart review.

\section*{Discussion}

Algorithmic recovery using clinically-driven roadmaps enhanced by LLM could offer a cost-effective, scalable way to reduce missingness in EHR data before developing or validating computable phenotypes, like the ALI. Using the LLM (context) expanded the search terms, while subsequent review by clinicians ensured that the final roadmap included only clinically-relevant ones (avoiding potential hallucinations). As proposed, the algorithm focuses on promoting completeness. Additional checks could be incorporated to address additional dimensions of EHR data quality, like correctness or plausibility.\cite{Weiskopf2013} If using an LLM that runs locally, such that no protected health information is shared, there are also exciting opportunities to enhance these quality checks using the LLM on EHR data directly.\cite{Ahsan2024,Ntinopoulos2025} 

We focused on operationalizing a well-studied, existing calculation for the ALI,\cite{Seeman2001} that takes $10$ component biomarkers, discretizes them into binary indicators of being unhealthy, and computes the ALI as the patient's proportion of unhealthy measurements. It assumes the same thresholds for all patients and equal weights for all components. Fortunately, our proposed algorithm could be used to develop alternative calculations or computable phenotypes for different outcomes. 

Our roadmap-based logic is straightforward and efficient to incorporate into EHR. SQL code can be written to retrieve ALI components, including labs, vitals, and the ICD-based surrogates developed in this paper. Optimal presentation of the ALI in EHR is a key next step, as ``alert fatigue'' is a well-known source of burnout for clinical staff.\cite{Ancker2017} Given the simplicity and interpretability of the ALI, there is an opportunity to present it to patients, perhaps through an EHR messaging portal such as MyChart.\cite{RAMSEY201829} This connection could eventually become bi-directional, wherein a patient uploads data from their personal devices (e.g., Oura ring) to their MyChart portal,\cite{Genes2018} and such information is incorporated in a future version of the ALI. 

Our choice of general-purpose LLM represents both a practical decision and an area for future methodological refinement. We chose to use Google Gemini's free tier for accessibility and cost-effectiveness. However, there are many alternatives, including ChatGPT,\cite{openai2023gpt4} Claude,\cite{claude3} Llama,\cite{llama3} and Copilot,\cite{microsoft2023copilot} each with capabilities that could influence search term generation quality. Our methodology only passed the roadmap structure to the LLM, rather than any patient data, which eliminates privacy concerns and allows flexibility in model selection without regulatory constraints. 
Training an LLM specifically on ICD-10 codes and medical terminology, rather than using a general-purpose model, could improve the validity and clinical relevance of proposed search terms, potentially reducing the noise inherent in broad keyword generation while enhancing the precision of diagnostic term matching for recovering missing data when biomarkers are not available. Future iterations would benefit from comparing performance across different LLM providers and exploring domain-specific model training to optimize the balance between comprehensive coverage and clinical accuracy in missing data recovery algorithms. 

There are many other worthwhile directions for future work. Inviting patients to identify gaps in their own EHR data (e.g., via survey) could provide valuable information about missingness mechanisms that would inform the roadmaps.\cite{Bell2020, Hagström2023, Haneuse2016} Our algorithm uses structured EHR data (ICD-10 codes), but text-mining unstructured fields (like free-text notes) could uncover additional sources of auxiliary information.\cite{SONABENDW2020104135} When re-fitting the healthcare utilization model, we used patients' ALIs from the algorithm with the LLM (context + clinicians) roadmap. Incorporating multiple versions of the same phenotype (e.g., ALI based on various roadmaps) instead could reduce bias and improve precision in statistical models.\cite{LU2024104690} If conducting some chart reviews is plausible, incorporating all three versions of a computable phenotype (extracted EHR data, chart reviews, and algorithm) could offer similar advantages in statistical analyses but would need to account for missing chart review data. 





\bibliographystyle{vancouver}
\bibliography{citation}





\section*{Acknowledgments}
The authors gratefully acknowledge Wake Forest University and Wake Forest University School of Medicine for an intercampus collaborative grant that supported the data collection for this work. 

\section*{Supplementary Materials}
\begin{itemize}
    \item \textbf{R code and additional tables and figures:} Available online at \url{https://github.com/LucyMcGowan/ehr-llm-validation/blob/main/supplement.pdf}.
\end{itemize}


\end{document}